%% file: main.tex
\documentclass[letterpaper, 10 pt, conference]{ieeeconf}  

\IEEEoverridecommandlockouts                              

\usepackage[T1]{fontenc}

\usepackage{amsfonts}	
\usepackage{amsmath}	
\usepackage{amssymb}    
\usepackage{siunitx}
\usepackage{xfrac}    
\usepackage{pifont}   

\usepackage{booktabs}
\usepackage{makecell}  
\usepackage[flushleft]{threeparttable}  
\usepackage{multirow}
\usepackage{rotating}

\usepackage{xspace}    
\usepackage[usenames,dvipsnames]{xcolor}    
\usepackage{colortbl}
\let\labelindent\relax
\usepackage[inline]{enumitem} 
\usepackage{graphicx} 
\usepackage{microtype}
\usepackage{cite}
\usepackage{flushend}

\makeatletter
\let\NAT@parse\undefined
\makeatother

\usepackage{url}

\usepackage[pdfencoding=auto, colorlinks=true]{hyperref} 
\usepackage[hang,flushmargin]{footmisc}

\usepackage[capitalize]{cleveref}
\crefname{section}{Sec.}{Secs.}
\Crefname{section}{Section}{Sections}
\Crefname{table}{Table}{Tables}
\crefname{table}{Tab.}{Tabs.}

\newcommand{\grayrule}{\arrayrulecolor{black!30}\midrule\arrayrulecolor{black}}

\newcommand{\cmark}{\ding{51}}  

\newcommand{\net}{\mbox{BEVCar}\xspace}

\begin{document}

\title{\LARGE \bf
BEVCar: Camera-Radar Fusion for BEV Map and Object Segmentation
}

\author{
Jonas Schramm$^{1*}$, 
Niclas Vödisch$^{1*}$, 
Kürsat Petek$^{1*}$, 
B Ravi Kiran$^{2}$, 
Senthil Yogamani$^{3}$, \\ 
Wolfram Burgard$^{4}$, 
and Abhinav Valada$^{1}$
\thanks{$^{*}$ Equal contribution.}%
\thanks{$^{1}$ Department of Computer Science, University of Freiburg, Germany.}%
\thanks{$^{2}$  Qualcomm SARL France.}%
\thanks{$^{3}$ QT Technologies Ireland Limited.}%
\thanks{$^{4}$ Department of Eng., University of Technology Nuremberg, Germany.}%
\thanks{This work was funded by Qualcomm Technologies Inc., the German Research Foundation (DFG) Emmy Noether Program grant No 468878300, and an academic grant from NVIDIA.}%
}

\maketitle
\thispagestyle{empty}
\pagestyle{empty}


\begin{abstract}
    \input{sections/0_abstract}
\end{abstract}


\input{sections/1_introduction}

\input{sections/2_related_work}

\input{sections/3_method}

\input{sections/4_experiments}
\input{sections/5_conclusion}


\footnotesize
\bibliographystyle{IEEEtran}
\bibliography{references.bib}


\end{document}

%% file: sections/0_abstract.tex
Semantic scene segmentation from a bird's-eye-view (BEV) perspective plays a crucial role in facilitating planning and decision-making for mobile robots. Although recent vision-only methods have demonstrated notable advancements in performance, they often struggle under adverse illumination conditions such as rain or nighttime. While active sensors offer a solution to this challenge, the prohibitively high cost of LiDARs remains a limiting factor. Fusing camera data with automotive radars poses a more inexpensive alternative but has received less attention in prior research. In this work, we aim to advance this promising avenue by introducing \net, a novel approach for joint BEV object and map segmentation. The core novelty of our approach lies in first learning a point-based encoding of raw radar data, which is then leveraged to efficiently initialize the lifting of image features into the BEV space. We perform extensive experiments on the nuScenes dataset and demonstrate that \net outperforms the current state of the art. Moreover, we show that incorporating radar information significantly enhances robustness in challenging environmental conditions and improves segmentation performance for distant objects. To foster future research, we provide the weather split of the nuScenes dataset used in our experiments, along with our code and trained models at \mbox{\url{http://bevcar.cs.uni-freiburg.de}}.

%% file: sections/1_introduction.tex
\section{Introduction}

Mobile robots such as autonomous vehicles heavily rely on accurate and robust perception of their environment. Therefore, robotic platforms are typically equipped with a variety of sensors~\cite{barnes2020oxford, caesar2020nuscenes, zheng2022tj4dradset}, each providing complementary information. For instance, surround-view cameras offer dense RGB images, while LiDAR or radar systems provide sparse depth measurements. However, fusing data from these different modalities poses a significant challenge due to inherently different data structures. A common approach to address this challenge is to employ a bird's-eye-view~(BEV) representation as a shared reference frame~\cite{li2022bevformer, liang2022bevfusion, harley2023simplebev, bai2022transfusion, man2023bevguide, kim2023crn}. 

While both LiDAR and radar data can be directly transformed into BEV space, camera-based information requires conversion from the image plane to a top-down view. Consequently, various lifting strategies have been proposed~\cite{philion2020lss, li2022bevformer, zhang2022beverse} resulting in tremendous performance improvements of vision-only approaches, some of which have been extended to incorporate LiDAR data~\cite{bai2022transfusion, liang2022bevfusion}. Despite the ability of LiDARs to yield highly accurate 3D point clouds, their suitability for large-scale deployment remains controversial due to their substantially higher costs compared to automotive radars. Nonetheless, camera-radar fusion has received considerably less attention from the research community, often only explored in addition to LiDAR input~\cite{hendy2020fishing, man2023bevguide}. In contrast, radar has been criticized as being too sparse to be effectively utilized in isolation~\cite{hendy2020fishing}.

\begin{figure}[t]
    \centering
    \includegraphics[width=\linewidth]{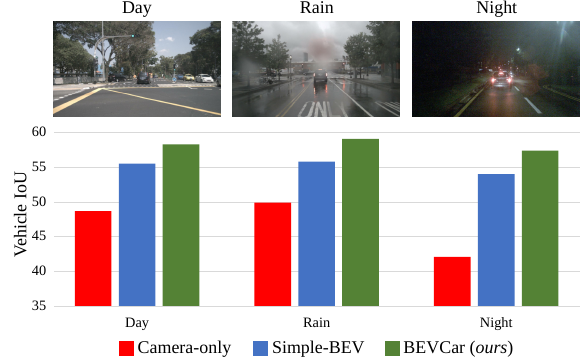}
    \vspace{-.6cm}
    \caption{We propose a novel method for BEV Camera-radar fusion (\net) for map and object segmentation. We demonstrate that \net yields more accurate predictions under adverse weather conditions than camera-only baselines while outperforming prior camera-radar works~\cite{harley2023simplebev}.}
    \label{fig:teaser}
\end{figure}

In this work, we underscore the pivotal role of radar in advancing robust robotic perception. Specifically, we focus on BEV object and map segmentation, highlighting the distinct advantage of radar in vision-impaired environmental conditions. While previous research has explored camera-radar fusion for BEV segmentation, some approaches necessitate additional LiDAR supervision during training~\cite{kim2023crn} or rely on specific radar metadata~\cite{harley2023simplebev, man2023bevguide}, which may not be accessible across models from different manufacturers.
To address these limitations, we propose a novel method that operates independently of such constraints. Our proposed \net architecture comprises two sensor-specific encoders and two attention-based modules for image lifting and BEV camera-radar fusion, respectively. Subsequently, we feed the fused features through a multi-task head to generate both map and object segmentation maps. We extensively evaluate our approach on the nuScenes~\cite{caesar2020nuscenes} dataset and demonstrate that it achieves state-of-the-art performance for camera-radar fusion while being robust in challenging illumination conditions.

The main contributions are as follows:
\begin{enumerate}[topsep=0pt]
    \item We introduce the novel \net for BEV map and object segmentation from camera and radar data.
    \item We propose a new attention-based image lifting scheme that exploits sparse radar points for query initialization.
    \item We show that learning-based radar encoding outperforms the usage of raw metadata.
    \item We extensively compare \net with previous baselines under challenging environmental conditions and demonstrate the advantage of utilizing radar measurements.
    \item We make the used day/night/rain splits on nuScenes~\cite{caesar2020nuscenes} publicly available and release our code and trained models at \mbox{\url{http://bevcar.cs.uni-freiburg.de}}.
\end{enumerate}

%% file: sections/2_related_work.tex
\section{Related Work}

In this section, we present an overview of vision-only methods operating in the bird's-eye-view (BEV) and review previous approaches for radar-based perception.


{\parskip=3pt
\noindent\textit{Camera-Based BEV Perception:} 
Current research in the field of camera-based BEV perception aims to handle the view discrepancy between the image space and the BEV space. Existing approaches typically employ an encoder-decoder architecture, incorporating a distinctive view transformation module to address spatial variations between the image and BEV planes.
Early works leverage variational autoencoders to decode features directly into a 2D top-view Cartesian coordinate system~\cite{lu2019ved}. In contrast, VPN~\cite{pan2020vpn} utilizes a multilayer perceptron~(MLP) to model dependencies across spatial locations in the image and BEV feature maps, ensuring global coverage in the view transformation.
Roddick~\textit{et~al.}~\cite{roddick2020pon} improve upon these works by introducing a more explicit geometry modeling. In particular, they propose a pyramid occupancy network with a per-scale dense transformer module to learn the mapping between a column in the image view and a ray in the BEV map. PoBEV~\cite{gosala2023pobev} extends this concept by processing flat and vertical features separately with distinct transformer modules resulting in further performance improvement.}

Recent methods can be categorized into lifting-based and attention-based mechanisms. Lifting-based approaches incorporate either an implicit depth distribution module~\cite{philion2020lss} to project features to a latent space or an explicit depth estimation module to generate an intermediate 3D output, e.g., for the tasks of object detection~\cite{you2019pseudolidar} or scene completion~\cite{li2023voxformer}. 
Attention-based approaches formulate view transformation as a sequence-to-sequence translation from the image space to BEV. TIIM~\cite{saha2022tiim} applies inter-plane attention between a polar ray in the BEV space and a vertical column in the image combined with self-attention across each respective polar ray with significant performance improvement with respect to depth-based approaches such as LSS~\cite{philion2020lss}. 

Recent advancements include full-surround view BEV perception approaches, such as CVT~\cite{zhou2022cvt} that uses a cross-view transformer with learned positional embeddings to avoid explicit geometric modeling and exploiting this BEV representation for policy learning~\cite{trumpp2023efficient}. In contrast, BEVFormer~\cite{li2022bevformer} and BEVSegFormer~\cite{peng2023bevsegformer} model geometry explicitly using camera calibration parameters and propose a deformable attention-based~\cite{zhu2020deformable} spatial cross-attention module for view unprojection. BEVFormer~\cite{li2022bevformer} additionally employs a temporal attention module for aggregating BEV maps over time using vehicle ego-motion, which represents the state of the art in 3D object detection. Temporal aggregation is also employed in BEVerse~\cite{zhang2022beverse}, which extends existing approaches with a motion prediction head and demonstrates that the proposed multi-task network outperforms single-task networks indicating a positive transfer among the tasks. The aforementioned approaches are often combined with novel data augmentation techniques~\cite{huang2021bevdet}, which address the view discrepancy between the image and BEV by maintaining spatial consistency across each intermediate embedding. Finally, SkyEye~\cite{gosala2023skyeye} proposes a less-constrained method that learns semantic BEV maps from labeled frontal view images by reconstructing semantic images over time. Our work leverages recent progress in monocular BEV perception and takes advantage of the radar modality for a more geometrically feasible view projection. This is achieved through a novel attention-based image lifting scheme using radar queries. Additionally, we propose to exploit existing image backbones that are pre-trained with contrastive learning to further regularize the modality-specific branches.
\looseness=-1


\begin{figure*}[t]
    \centering
    \includegraphics[width=\linewidth]{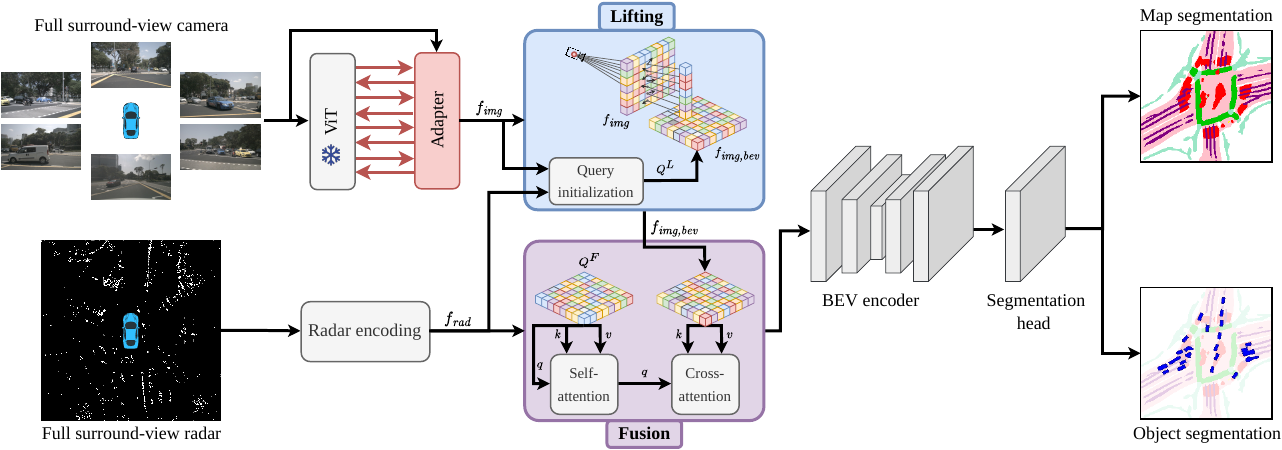}
    \vspace*{-.6cm}
    \caption{Overview of our proposed \net approach for camera-radar fusion for BEV map and object segmentation. We utilize a frozen DINOv2~\cite{oquab2023dinov2} with a learnable adapter to encode the surround-view images. Inspired by LiDAR-based perception~\cite{zhou2018voxelnet}, we employ a learnable radar encoding instead of processing the raw metadata. We then lift the image features to the BEV space via deformable attention including the novel radar-driven query initialization scheme. Finally, we fuse the lifted image representation with the learned radar features in an attention-based manner and perform multi-class BEV segmentation for both vehicles and the map categories.}
    \label{fig:overview}
    \vspace*{-.3cm}
\end{figure*}


{\parskip=3pt
\noindent\textit{Radar-Based Perception:}
Radars measure the distance to a target based on the time difference between emitting a radio wave and receiving its reflection. Published datasets for robotic applications include different types of radars such as spinning radars~\cite{barnes2020oxford}, automotive radars~\cite{caesar2020nuscenes}, or 4D imaging radars~\cite{zheng2022tj4dradset}. In this work, we focus on automotive radars.
As radar poses a comparably inexpensive technology to measure distance directly, it has been leveraged to improve vision-based 3D object detection. While ClusterFusion~\cite{kurniawan2023clusterfusion} merges radar and camera data only in the image space, SparseFusion3D~\cite{yu2023sparsefusion3d} performs sensor fusion both in the image and BEV space.

In segmentation, initial works investigated semantic segmentation of radar point clouds~\cite{schumann2018semantic} without complementary vision input.
More recently, research towards multi-modal BEV map and object segmentation has received growing attention. The authors of the pioneering work FISHING~Net~\cite{hendy2020fishing} propose an MLP-based lifting strategy for camera features. To combine these features with radar data, which are encoded by a UNet-like network, FISHING~Net performs class-based priority pooling.
In contrast, \mbox{Simple-BEV}~\cite{harley2023simplebev} processes the raw radar data in a rasterized BEV format and concatenates these with image features that are lifted via bilinear sampling. Although \mbox{Simple-BEV} targets object-agnostic vehicle segmentation, the training relies on additional instance information for object center and pixel offset prediction.
Since purely concatenation-based fusion might suffer from spatial misalignment, CRN~\cite{kim2023crn} employs deformable attention~\cite{zhu2020deformable} to aggregate image and radar features. However, the method uses LSS~\cite{philion2020lss} for lifting the image features and requires LiDAR during training to supervise the depth distribution network.
Finally, BEVGuide~\cite{man2023bevguide} does not exploit further knowledge other than available during deployment. Using homography-based projection, features from the EfficientNet~\cite{tan2019efficientnet} image backbone are transformed into a scale-ambiguous top-down representation. The radar data is converted to BEV space and then encoded by two convolutional layers. In contrast to prior works, BEVGuide proposes a bottom-up lifting approach by querying the sensor features from a unified BEV space to obtain sensor-specific embeddings that are then concatenated.
In this work, we further advance these ideas and utilize a more refined radar encoder that is inspired by LiDAR processing~\cite{zhou2018voxelnet}. Moreover, we propose a novel lifting scheme that explicitly leverages radar points as a strong prior.
}

%% file: sections/3_method.tex
\section{Technical Approach}

In this section, we present our proposed \net approach for BEV object and map segmentation from surround-view cameras and automotive radar.
As illustrated in \cref{fig:overview}, \net comprises two sensor-specific encoders for image and radar data, respectively. We lift the image features to the BEV space via deformable attention, where we utilize radar data to initialize the queries. Following an intermediate fusion strategy, we then combine the lifted image representation with the learned radar features using a cross-attention module. Finally, we reduce the spatial resolution in a bottleneck operation and perform BEV segmentation for both vehicles and the map with a single multi-class head.
We provide further details of each step in the following subsections.


\subsection{Sensor Data Encoding}
\label{ssec:sensor-encoding}

As depicted in \cref{fig:overview}, we process the raw data of both modalities in two separate encoders.

{\parskip=3pt
\noindent\textit{Camera:} 
For encoding the camera data, we employ a frozen DINOv2 ViT-B/14~\cite{oquab2023dinov2}, whose image representation captures more semantic information than ResNet-based backbones~\cite{he2016resnet}. Following the common approach~\cite{kaeppeler2024spino, Xu2023san}, we utilize a ViT~adapter~\cite{chen2022vision} with learnable weights. To cover the surround-view vision, we concatenate the images from $N$ cameras at each timestamp resulting in an input dimension of $N \times H \times W$, where $H$ and $W$ denote the image height and width, respectively. For downstream processing, the ViT~adapter outputs multi-scale feature maps with $F$ channels that correspond to scales \sfrac{1}{4}, \sfrac{1}{8}, \sfrac{1}{16}, and \sfrac{1}{32} of the image size.
}

\begin{figure}[t]
    \centering
    \includegraphics[width=\linewidth]{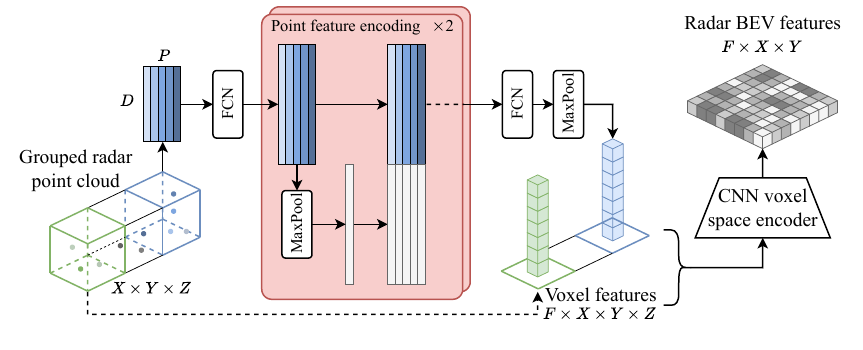}
    \vspace{-.8cm}
    \caption{Inspired by LiDAR processing, we encode the radar data with fully connected layers (FCN) in a point-wise manner and combine point features within a voxel with max pooling. Subsequently, we employ a CNN-based height compression to obtain the overall radar features in the BEV space.}
    \label{fig:radar-encoding}
\end{figure}

{\parskip=3pt
\noindent\textit{Radar:} 
The radar data is represented by a point cloud with various features available for each point. Unlike prior works~\cite{harley2023simplebev, man2023bevguide}, we emphasize that relying on the built-in post-processing from a specific radar model makes a method less versatile. Hence, similar to SparseFusion3D~\cite{yu2023sparsefusion3d}, we utilize only $D$ basic characteristics of a radar point: the 3D position $(x, y, z)$, uncompensated velocities $(v_x, v_y)$, and the radar cross-section $\textit{RCS}$, which captures the detectability of a surface. Instead of utilizing the raw data~\cite{harley2023simplebev}, we propose to learn a radar representation inspired by encoding LiDAR point clouds~\cite{zhou2018voxelnet}.
First, we group the radar points based on their spatial position in a voxel grid of size $X \times Y \times Z$ that corresponds to the resolution of the BEV space and a discretization in height. To restrict memory requirements and alleviate bias towards high-density voxels, we employ random sampling in those voxels that contain more than $P$ radar points. Each point including its metadata is then fed through the point feature encoding as illustrated in \cref{fig:radar-encoding}, where FCN refers to fully connected layers. Note that the point feature encoding does not accumulate information from multiple voxels. Subsequently, we employ max pooling for each voxel to obtain a single feature vector of size $F$. Finally, we feed the voxel features through a CNN-based voxel space encoder to compress the features along the height dimension, resulting in the overall radar BEV encoding~$f_\mathit{rad}$.
}

\subsection{Image Feature Lifting}

We follow a learning-based approach to lift the encoded vision features from the 2D image plane to the BEV space. Inspired by BEVFormer~\cite{li2022bevformer}, we utilize deformable attention~\cite{zhu2020deformable} but propose a novel query initialization scheme that exploits sparse radar points.

\begin{figure}
    \centering
    \includegraphics[width=\linewidth]{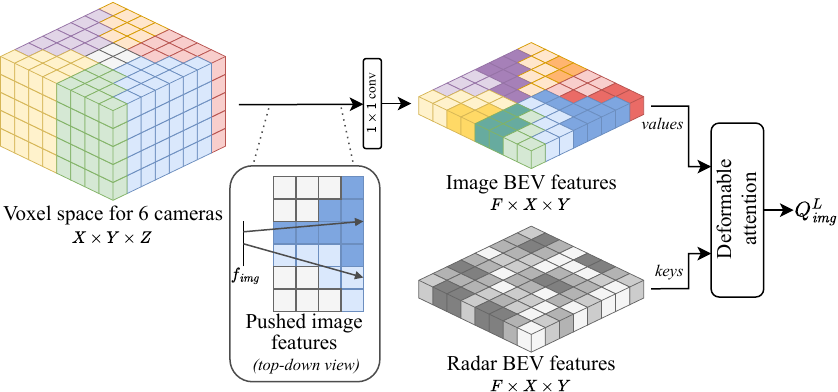}
    \caption{Our data-driven query initialization scheme leverages 3D radar information to guide lifting the 2D image features to the BEV space. While the image BEV features are only obtained from uniform assignment along camera rays, the final $Q_\mathit{img}^L$ considers depth from radar via deformable attention.}
    \label{fig:query-initialization}
    \vspace{-.3cm}
\end{figure}

{\parskip=3pt
\noindent\textit{Query Initialization:} 
The core motivation of our proposed query initialization scheme is to leverage 3D information from radar measurements for an initial lifting step of the 2D image features to the BEV space.
As visualized in \cref{fig:query-initialization}, we first create a voxel space of size $X \times Y \times Z$ that is defined by the BEV resolution $X \times Y$, an additional height discretization~$Z$, and centered at the forward-facing camera.
Second, we assign each voxel to one or two cameras based on their fields of view.
Third, we push the vision features from the 2D image plane to the 3D voxel space via ray projection, i.e., each voxel within a frustum along a ray contains the same image features. In particular, we utilize the image features from scale \sfrac{1}{8}. If the fields of view of two cameras overlap, we average the features in the affected voxels.
Subsequently, we employ a $1 \times 1$ convolutional layer to remove the height component resulting in an $X \times Y$ voxel grid with $F$ feature channels.
Note that at this stage, the image features are still uniformly distributed without a notion of depth. Therefore, we use deformable attention~\cite{zhu2020deformable} guided by the sparse radar point cloud to filter the feature map resulting in the initialized query $Q_\mathit{img}^L$ of size $F \times X \times Y$.
}

{\parskip=3pt
\noindent\textit{Lifting:}
In the next step, we combine our data-driven initial queries~$Q_\mathit{img}^L$ with a learnable position embedding~$Q_\mathit{pos}^L$ to achieve permutation invariance and learnable BEV queries~$Q_\mathit{bev}^L$~\cite{li2022bevformer, harley2023simplebev}:
\begin{equation}
    Q^L = Q_\mathit{img}^L + Q_\mathit{pos}^L + Q_\mathit{bev}^L
    \label{eqn:lifting-queries}
\end{equation}
Employing deformable attention~\cite{zhu2020deformable} again, we construct a 3D voxel space of size $X \times Y \times Z$ to pull the vision encoding from the images. In contrast to the query initialization, we now sample offsets on the image planes instead of the BEV space.
After six cascaded transformer blocks, we obtain the final feature map $f_\mathit{img, bev}$ that has the same dimension as the encoded radar data, i.e., $F \times X \times Y$.
}

\subsection{Sensor Fusion}

For fusing the lifted image features with the encoded radar data, we follow a scheme comparable to the lifting step. Inspired by TransFusion~\cite{bai2022transfusion}, which fuses camera and LiDAR for 3D object detection, we query image features in the surroundings of the radar points and extract the values via deformable attention~\cite{zhu2020deformable}. Similar to \cref{eqn:lifting-queries}, we form the initial query by summing the encoded radar data~$f_\mathit{rad}$, a learnable position encoding~$Q_\mathit{pos}^F$, and learnable BEV queries~$Q_\mathit{bev}^F$:
\begin{equation}
    Q^F = f_\mathit{rad} + Q_\mathit{pos}^F + Q_\mathit{bev}^F
    \label{eqn:fusion-queries}
\end{equation}
Importantly, the lifted image features only serve as keys and values during the cross-attention step. In total, we utilize a cascade of six transformer blocks.
Finally, to encode the features of both modalities in a joint manner, we feed the output of the last block through a ResNet-18~\cite{he2016resnet} bottleneck, referred to as BEV encoder in \cref{fig:overview}.


\begin{table*}[t]
\footnotesize
\centering
\caption{Baseline Comparison on the nuScenes dataset}
\vspace{-0.2cm}
\label{tab:baselines}
\setlength\tabcolsep{12.0pt}
\begin{threeparttable}
    \begin{tabular}{ l | cc | ccccc }
        \toprule
        \textbf{Method} & \textbf{Modalities} & \textbf{Image Backbone} & Vehicle & Driv. Area & Lane & Map & mIoU \\
        \midrule
        CVT~\cite{zhou2022cvt} & C & EfficientNet & 36.0 & 74.3 & -- & -- & 55.2 \\ 
        BEVFormer-S~\cite{li2022bevformer} & C & ResNet-101 & 43.2 & 80.7 & 21.3 & -- & 62.0 \\
        Simple-BEV~\cite{harley2023simplebev} & C & ResNet-101 & 47.4 & -- & -- & -- & -- \\
        \grayrule
        Simple-BEV~\cite{harley2023simplebev} & C+R & ResNet-101 & 55.7 & -- & -- & -- & -- \\
        Simple-BEV++ & C+R & ResNet-101 & 52.7 & 77.7 & 35.8 & 46.1 & 65.2 \\ 
        Simple-BEV++ & C+R & ViT-B/14 & 54.5 & 81.2 & 40.4 & 50.4 & 67.9 \\
        BEVGuide~\cite{man2023bevguide} & C+R & EfficientNet & \textbf{59.2} & 76.7 & \underline{44.2} & -- & 68.0 \\
        CRN~\cite{kim2023crn} & C+R (+L) & ResNet-50 & \underline{58.8} & \underline{82.1} & -- & -- & \underline{70.5} \\
        \grayrule
        BEVCar (\textit{camera}) & C & ViT-B/14 &  48.8 & 81.1 & 40.6 & 50.5 & 65.0 \\
        \net (\textit{ResNet}) & C+R & ResNet-101 & 57.3 & 81.8 & 43.8 & \underline{53.0} & 69.6 \\
        \net (\textit{ours}) & C+R & ViT-B/14 & 58.4 & \textbf{83.3} & \textbf{45.3} & \textbf{54.3} & \textbf{70.9} \\
        \bottomrule
    \end{tabular}
    \footnotesize
    We compare \net with both camera-only (C) and camera-radar (C+R) BEV segmentation methods on the nuScenes~\cite{caesar2020nuscenes} validation split.
    \mbox{Simple-BEV++} is a customized \mbox{Simple-BEV}~\cite{harley2023simplebev} without instance-aware training but with the same radar metadata and map segmentation head as our method.
    Note that CRN~\cite{kim2023crn} uses LiDAR during training.
    The ``map'' metric averages the IoU of all nuScenes map classes.
    Previous works report predictions for fewer classes, indicated by ``--''. To compare \net with these methods, we provide the mean of ``vehicle'' and ``drivable area'' classes as ``mIoU''.
    Bold and underlined values denote the best and second-best metrics per column, respectively.
\end{threeparttable}
\vspace*{-.4cm}
\end{table*}


\subsection{Segmentation Head}
\label{ssec:segmentation-head}

We employ a single head for multi-class BEV semantic segmentation. In detail, we utilize two convolutional layers with ReLU activations followed by a final $1\times1$ convolutional layer to output one object class and $M$ map classes. Given the BEV space resolution, the segmentation head produces an output of size $(M+1) \times X \times Y$. Thus, a pixel can not only capture both a vehicle and a map class prediction but can also be assigned to multiple map categories.

{\parskip=3pt
\noindent\textit{Object Segmentation:}
For segmenting objects, we consider all vehicle-like entities, e.g., passenger cars and trucks. Unlike prior works~\cite{harley2023simplebev}, we emphasize that object-agnostic segmentation should not rely on instance-aware information during training time as this renders the application of a method less flexible due to requiring additional annotations. Therefore, we supervise the object channel of the segmentation head solely via the binary cross-entropy loss:
\begin{equation}
    \mathcal{L}_\mathit{BCE} = \frac{-1}{N} \sum_{i=1}^N \log (p_{i,t}) \, ,
\end{equation}
where $p_{i,t}$ is defined per pixel $i \in [1, N]$ as:
\begin{equation}
    p_{i,t} = 
    \begin{cases}
    p_i & \text{if } y_i = 1 \\
    1 - p_i & \text{otherwise.}
    \end{cases}
\label{eqn:p-it}
\end{equation}
The binary ground truth label $y_i \in \{0, 1\}$ specifies whether the pixel $i$ belongs to the vehicle class. The corresponding predicted probability for $y_i = 1$ is denoted by $p_i$.
}

{\parskip=3pt
\noindent\textit{Map Segmentation:}
While most previous methods~\cite{li2022bevformer, man2023bevguide, kim2023crn} predict only the road and occasionally also lane dividers, we include further map classes such as pedestrian crossings and walkways. For an exhaustive list, please refer to \cref{ssec:experimental-settings}. To supervise the map channels of the segmentation head during training, we employ a multi-class variant of the \mbox{$\alpha$-balanced} focal loss~\cite{lin2020focal}:
\begin{equation}
    \mathcal{L}_\mathit{FOC} = \sum_{c=1}^C \frac{-1}{N} \sum_{i=1}^N \alpha_{i,t} \left( 1 - p_{i,t} \right)^\gamma \log (p_{i,t}) \, ,
    \label{eqn:focal-loss}
\end{equation}
where $c \in [1, C]$ refers to the semantic classes and $\gamma$ is a focusing parameter to differentiate between easy/hard examples. Additionally, $\alpha_{i,t}$ is defined analogously to \cref{eqn:p-it}:
\begin{equation}
    \alpha_{i,t} =
    \begin{cases}
        \alpha & \text{if } y_i = 1 \\
        1 - \alpha & \text{otherwise,}
    \end{cases}
\end{equation}
with tunable parameter $\alpha$ to address the foreground-background imbalance.
}

%% file: sections/4_experiments.tex
\section{Experimental Evaluation}

In this section, we outline the experimental setup and compare our \net approach to various baselines. We further analyze the impact of the components of our method and demonstrate the advantage of radar measurements over vision-only methods under adverse conditions.


\subsection{Experimental Settings}
\label{ssec:experimental-settings}

We introduce the utilized dataset and metrics for evaluation and provide further implementation details.

{\parskip=3pt
\noindent\textit{Dataset and Metrics:} 
We evaluate our \net approach on the nuScenes dataset~\cite{caesar2020nuscenes} for automated urban driving in Singapore and Boston, MA, being the only publicly available dataset that provides the required sensor data and ground truth map annotations. The nuScenes dataset comprises surround-view vision from six RGB cameras and five automotive radars and provides BEV map information. For training and evaluation, we use the official training/validation split, containing 28,130 and 6,019 samples, respectively. We further categorize the validation scenes into day (4,449 samples), rain (968 samples), and night (602 samples) scenes and include this split in our code release.  For object segmentation, we combine all subclasses of the ``vehicle'' category. For map segmentation, we consider all available classes, i.e., ``drivable area'', ``carpark area'', ``pedestrian crossing'', ``walkway'', ``stop line'', ``road divider'', and ``lane divider''.
We report individual intersection over union (IoU) metrics~\cite{hurtado2022semantic} for those classes that have been addressed by prior works and refer to the mean IoU of all map classes by ``map''. To compare \net with previous baselines that predict fewer classes, we report the average of ``vehicle'' and ``drivable area'' as ``mIoU''.
}

{\parskip=3pt
\noindent\textit{Implementation Details:} 
Similar to related work~\cite{harley2023simplebev, man2023bevguide, kim2023crn}, our BEV grid covers an area of \SI{100}{\meter}$\times$\SI{100}{\meter} centered at the ego vehicle and is discretized at a resolution of $200\times200$ cells. We further construct an up/down span from the ground to \SI{10}{\meter} in height and discretize it into eight bins. The resulting 3D tensor is oriented with respect to the forward-facing camera serving as the reference coordinate system.
For both training and inference, we resize the images of the six cameras to $448\times896$ pixels adapting the results of an analysis from Harley~\textit{et~al.}~\cite{harley2023simplebev} to the requirements of the employed ViT adapter. In accordance with the released code of the same study, we aggregate five radar sweeps as input.
During training, we set the parameters of the focal loss (see \cref{eqn:focal-loss}) to $\alpha = 0.25$ and $\gamma=3$.
}

\subsection{Quantitative Results}

We compare \net to various baseline works in \cref{tab:baselines}, including the camera-radar fusion methods \mbox{Simple-BEV}~\cite{harley2023simplebev}, BEVGuide~\cite{man2023bevguide}, and CRN~\cite{kim2023crn}, which leverages depth from LiDAR during training. At the time of submission, only the authors of \mbox{Simple-BEV} released their code. We utilize this for an extended version \mbox{Simple-BEV++} by adding the BEV map segmentation task, removing additional radar metadata (see \cref{ssec:sensor-encoding}), and disregarding the instance-aware losses (see \cref{ssec:segmentation-head}). To demonstrate the advantage of radar measurements, we further compare \net to the vision-only baselines CVT~\cite{zhou2022cvt}, BEVFormer~\cite{li2022bevformer}, and variations of both \mbox{Simple-BEV}~\cite{harley2023simplebev} and our proposed \net.

Concerning the latter, our camera-only version of \net yields a small increase of performance over the \mbox{Simple-BEV}~(C) baseline for the ``vehicle'' class ($+1.4$~IoU) and over the static version of BEVFormer for the ``drivable area'' class ($+0.4$~IoU). We primarily attribute the improvement within the vision-only regime to the semantically rich image representation of the DINOv2~\cite{oquab2023dinov2} backbone. Integrating radar data via our proposed methodology results in substantially enhanced vehicle predictions ($+9.6$~IoU) and noteworthy improvements in map segmentation ($+3.8$~mIoU). We thus infer that utilizing radar for robotic perception promises significantly better performance and further analyze this claim in \cref{ssec:ablations} under various aspects. 

For the vehicle segmentation task, \net outperforms \mbox{Simple-BEV}~($+2.7$~IoU) and achieves comparable performance to BEVGuide~($-0.8$~IoU) and CRN~($-0.4$~IoU). Concerning CRN, it is important to consider that this method relies on LiDAR during the training phase to learn metric depth. For map segmentation, \net improves upon all baselines while providing information for more semantic classes. With respect to the combined evaluation for both tasks, \net achieves the highest performance across the board with $+2.9$~mIoU versus BEVGuide and $+0.4$~mIoU versus CRN. We further compare \net to the aforementioned \mbox{Simple-BEV++}. To eliminate the impact of different backbones, we integrate both \mbox{ResNet-101}~\cite{he2016resnet} and DINOv2 \mbox{ViT-B/14}~\cite{oquab2023dinov2} in either method. Note that the multi-task training of \mbox{Simple-BEV++} leads to reduced performance for vehicle segmentation over the \mbox{Simple-BEV} baseline. Although we observe that the DINOv2 backbone also improves the results of \mbox{Simple-BEV++}, our \net approach still outperforms \mbox{Simple-BEV++} with both image backbones \mbox{ResNet-101}~($+4.4$~mIoU) and \mbox{ViT-B/14}~($+3.0$~mIoU), demonstrating the novelty of our method.

In \cref{fig:results}, we underline this observation by visualizing the improvements and errors of \net compared to \mbox{Simple-BEV++}. We further show the ground truth BEV object and map segmentation and provide visual predictions from the camera-only baseline, \mbox{Simple-BEV++}, and our \net approach. For a detailed analysis of the different weather and illumination conditions, please refer to the next section.


\begin{table}[t]
\footnotesize
\centering
\caption{Components Analysis}
\vspace{-0.2cm}
\label{tab:ablation-components}
\setlength\tabcolsep{8.0pt}
\begin{threeparttable}
    \begin{tabular}{ l | c c }
        \toprule
        \textbf{Method} & Vehicle & Map \\
        \midrule
        \multicolumn{2}{l}{\textit{Radar encoding}} \\
        [.25ex]
        No radar encoding~\cite{harley2023simplebev} & 57.8 & 53.4 \\
        \net (\textit{ours}) & 58.4 & 54.3 \\
         & (\textit{+0.6}) & (\textit{+0.9}) \\
        \grayrule
        \multicolumn{2}{l}{\textit{Lifting and fusion}} \\
        [.25ex]
        Parameter-free~\cite{harley2023simplebev} & 56.6 & 50.1 \\
        \net (\textit{ours}) & 58.4 & 54.3 \\
        & (\textit{+1.8}) & (\textit{+4.2}) \\
        \bottomrule
    \end{tabular}
    \footnotesize
    We demonstrate the efficacy of our employed radar encoding and our attention-based lifting and fusion scheme compared to simpler approaches.
    The ``map'' metric denotes the mean IoU of all map classes.
\end{threeparttable}
\vspace{-.4cm}
\end{table}


\subsection{Ablations and Analysis}
\label{ssec:ablations}

To further analyze our proposed \net approach, we provide ablations for its components and evaluate its performance under challenging conditions.

{\parskip=3pt
\noindent\textit{Components Analysis:}
We evaluate the impact of two key components of \net, i.e., the proposed radar point encoding and the new radar-driven image feature lifting, and report the improvements over baselines inspired by \mbox{Simple-BEV}~\cite{harley2023simplebev} in \cref{tab:ablation-components}.
First, compared to utilizing the raw radar data without a learning-based encoding, our approach yields $+0.6$~IoU and $+0.9$~mIoU for the vehicle and map segmentation tasks, respectively. Second, while the baseline uses a parameter-free lifting of the image features to the BEV space, our attention-based scheme leverages radar information already during the lifting stage. In comparison, this results in an increase of $+1.8$~IoU for vehicle segmentation and $+4.2$~mIoU for map segmentation.
}

{\parskip=3pt
\noindent\textit{Distance-Based Object Segmentation:} 
In \cref{tab:perception-range}, we analyze the vehicle segmentation quality of \net, its camera-only variant, \mbox{Simple-BEV}~\cite{harley2023simplebev}, and \mbox{Simple-BEV++} for three different range intervals including $0$-$\SI{20}{\meter}$, $20$-$\SI{35}{\meter}$, and \mbox{$35$-$\SI{50}{\meter}$}. Note that the overall performance of \mbox{Simple-BEV} is slightly lower than reported in \cref{tab:baselines} due to rerunning the authors' code to enable the range-based evaluation.
Generally, we observe that the results of the camera-only baseline significantly differ between the evaluation criteria. While the IoU in the $0$-$\SI{20}{\meter}$ range is comparable to \mbox{Simple-BEV}, for the $35$-$\SI{50}{\meter}$ range it achieves only half of the initial performance. Although the general trend is similar among all camera-radar methods, the effect is the least severe for \net. Our experiment demonstrates the advantage of utilizing radar measurements to maintain object segmentation performance also at larger distances.
}


\begin{table}[t]
\footnotesize
\centering
\caption{Vehicle Perception Range}
\vspace{-0.2cm}
\label{tab:perception-range}
\setlength\tabcolsep{3.0pt}
\begin{threeparttable}
    \begin{tabular}{ l | c | c !{\color{gray}\vline} ccc }
        \toprule
        & & & \multicolumn{3}{c}{Range Intervals} \\
        \textbf{Method} & \textbf{Modalities} & $0$-$\SI{50}{\meter}$ & $0$-$\SI{20}{\meter}$ & $20$-$\SI{35}{\meter}$ & $35$-$\SI{50}{\meter}$ \\
        \midrule
        BEVCar (\textit{camera}) & C & 48.8 & 68.7 & 45.8 & 29.1 \\
        Simple-BEV~\cite{harley2023simplebev} & C+R & \underline{55.5} & 70.1 & \underline{53.6} & \underline{39.1} \\
        Simple-BEV++ & C+R & 54.5 & \underline{71.4} & 52.4 & 36.6 \\
        \net (\textit{ours}) & C+R & \textbf{58.4} & \textbf{74.0} & \textbf{56.5} & \textbf{41.0} \\
        \bottomrule
    \end{tabular}
    \footnotesize
    The modalities denote camera (C) and radar (R) input.
    \mbox{Simple-BEV++} and BEVCar use a \mbox{ViT-B/14} image backbone. For \mbox{Simple-BEV}, we utilize the model trained by the authors.
\end{threeparttable}
\end{table}

\begin{table}[t]
\footnotesize
\centering
\caption{Weather and Illumination Analysis}
\vspace{-0.2cm}
\label{tab:results-weather}
\setlength\tabcolsep{4.75pt}
\begin{threeparttable}
    \begin{tabular}{ l | cc !{\color{gray}\vline} cc !{\color{gray}\vline} cc }
        \toprule
        & \multicolumn{2}{c !{\color{gray}\vline}}{Day} & \multicolumn{2}{c !{\color{gray}\vline}}{Rain} & \multicolumn{2}{c}{Night} \\
        \textbf{Method} & Vehicle & Map & Vehicle & Map & Vehicle & Map \\
        \midrule
        BEVCar (\textit{camera}) & 48.7 & \underline{53.3} & 49.9 & 45.1 & 42.1 & \underline{39.2} \\
        Simple-BEV~\cite{harley2023simplebev} & \underline{55.5} & -- & \underline{55.8} & -- & \underline{54.0} & -- \\
        Simple-BEV++ & 54.4 & 53.1 & 54.8 & \underline{45.6} & 52.4 & 37.8 \\
        \net (\textit{ours}) & \textbf{58.3} & \textbf{57.3} & \textbf{59.1} & \textbf{48.8} & \textbf{57.4} & \textbf{42.4} \\
        \bottomrule
    \end{tabular}
    \footnotesize
    ``Map'' denotes the mean IoU of all map classes. \mbox{Simple-BEV++} and BEVCar use a \mbox{ViT-B/14} image backbone. For \mbox{Simple-BEV}, we utilize the model trained by the authors, which does not perform map segmentation.
\end{threeparttable}
\vspace{-.4cm}
\end{table}

\begin{table}[t]
\footnotesize
\centering
\caption{Runtime Analysis}
\vspace{-0.2cm}
\label{tab:runtime}
\setlength\tabcolsep{6pt}
\begin{threeparttable}
    \begin{tabular}{ l | c | c | c !{\color{gray}\vline} c }
        \toprule
        & \textbf{Image} & \textbf{Deformable} & \multicolumn{2}{c}{Runtime} \\
        \textbf{Method} & \textbf{Backbone} & \textbf{Attention} & ms & FPS \\
        \midrule
        Simple-BEV~\cite{harley2023simplebev} & ResNet-101 & & 18 & 54.9 \\
        \net & ResNet-101 & \cmark & 137 & 7.3 \\
        \grayrule
        BEVCar (\textit{camera}) & ViT-B/14 & \cmark & 352 & 2.8 \\
        Simple-BEV++ & ViT-B/14 & \cmark & 277 & 3.6 \\
        \net (\textit{ours}) & ViT-B/14 & \cmark & 382 & 2.6 \\
        \bottomrule
    \end{tabular}
    \footnotesize
    Runtime of a forward pass measured on an Nvidia A100 GPU.
\end{threeparttable}
\vspace{-.5cm}
\end{table}

\begin{figure*}[t]
    \centering
    \includegraphics[width=\linewidth]{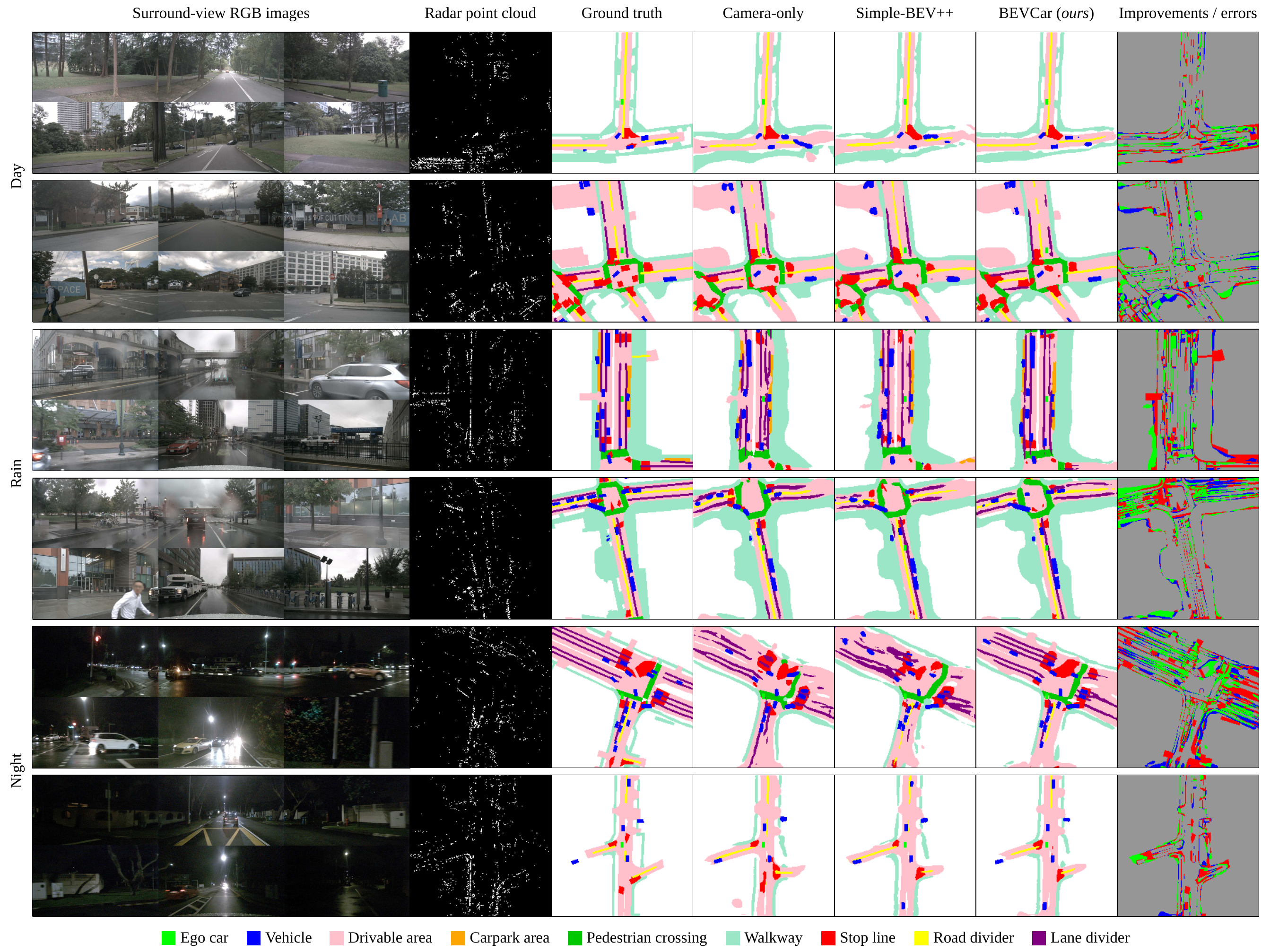}
    \vspace*{-.6cm}
    \caption{Qualitative results of our proposed \net, the camera-only baseline, and \mbox{Simple-BEV++} (\mbox{ViT-B/14}), for which we also show the improvement/error map. Pixels misclassified by \mbox{Simple-BEV++} and correctly predicted by \net are shown in green, pixels misclassified by \net and correctly predicted by \mbox{Simple-BEV++} in blue, and pixels misclassified by both models in red.}
    \label{fig:results}
    \vspace*{-.2cm}
\end{figure*}


{\parskip=3pt
\noindent\textit{Robustness to Weather and Illumination:}
Besides providing complementary information, i.e., dense RGB data versus sparse distance and velocity measurements, a core difference between cameras and radars is the source of energy utilized by the respective sensor. While passive sensors such as cameras rely on an external source like the sun, active sensors such as radars provide their own energy. Therefore, passive sensors suffer from challenging illumination conditions, e.g., faced during rain or at night. We thus emphasize that evaluating automotive perception systems specifically in these situations is imperative to understand their performance fully.

In \cref{tab:results-weather}, \cref{fig:teaser}, and \cref{fig:results}, we separate the previously reported metrics for \net and the same baselines as in the study on the perception range into \textit{day}, \textit{rain}, and \textit{night}. We observe that the vehicle segmentation IoU of the camera-only baseline is subject to substantial degradation during the night. In contrast, all camera-radar methods can maintain their performance, whereas \net achieves the highest performance. On the other hand, the map segmentation mIoU decreases during rain and even further at night, which holds for all investigated methods. The results indicate that radar is most beneficial for object detection and less relevant for BEV mapping, which is expected as depth information is less important for learning a mapping of the planar map classes from the 2D image space to the BEV space than mapping objects with defined height, width, and depth parameters.
}

{\parskip=3pt
\noindent\textit{Runtime Analysis:}
In \cref{tab:runtime}, we report the runtime of our proposed \net and multiple baseline methods, measured on an Nvidia A100 GPU and averaged over the validation split. In contrast to Harley~\textit{et~al.}~\cite{harley2023simplebev}, we only consider the forward pass without data loading and loss calculation. Most notable is the slow-down caused by deformable attention~\cite{zhu2020deformable} employed in our proposed lifting module as well as in the ViT adapter. Importantly, in comparison to the vision-only baseline, including the radar information does not result in a significantly higher runtime. Note that the frequency of the synchronized keyframes in the nuScenes~\cite{caesar2020nuscenes} dataset is \SI{2}{\hertz}.
}

%% file: sections/5_conclusion.tex
\section{Conclusion}

In this work, we introduced \net addressing camera-radar fusion for BEV map and object segmentation.
\net comprises a new learning-based radar point encoding and leverages radar information early during the lifting step of the vision features from the image plane to the BEV space.
We demonstrated that \net outperforms previous camera-radar approaches when jointly considering map and object segmentation.
We extensively evaluated the performance in challenging weather and illumination conditions and analyzed the robustness for various perception ranges. Our results clearly demonstrate the benefit of utilizing automotive radar in addition to surround-view vision.
To facilitate further research in this direction, we include our day/rain/night split of the nuScenes~\cite{caesar2020nuscenes} validation data in the public release of our code. In the future, we will address robustness in case of partial or complete sensor failure, e.g., by leveraging cross-modality distillation during training of the network.